\documentclass[runningheads]{llncs}
\usepackage{multirow}
\usepackage{graphicx}
\usepackage{amsmath}
\begin{document}
\title{Explainable concept mappings of MRI: Revealing the mechanisms underlying deep learning-based brain disease classification}
\titlerunning{Explainable concept mappings of MRI}
%
\author{Christian Tinauer\inst{1}\orcidID{0000-0003-4355-3898} \and
Anna Damulina\inst{1}\orcidID{0000-0003-4629-6380} \and
Maximilian Sackl\inst{1}\orcidID{0000-0002-1927-7953} \and
Martin Soellradl\inst{2}\orcidID{0000-0002-3760-5578} \and
Reduan Achtibat\inst{3}\orcidID{0009-0008-0052-3836} \and
Maximilian Dreyer\inst{3}\orcidID{0009-0007-9069-6265} \and
Frederik Pahde\inst{3}\orcidID{0000-0002-5681-6231} \and
Sebastian Lapuschkin\inst{3}\orcidID{0000-0002-0762-7258} \and
Reinhold Schmidt\inst{1}\orcidID{0000-0002-6406-7584} \and
Stefan Ropele\inst{1}\orcidID{0000-0002-5559-768X} \and
Wojciech Samek\inst{3,4,5}\orcidID{0000-0002-6283-3265} \and
Christian Langkammer\inst{1}\orcidID{0000-0002-7097-9707}}
\authorrunning{Tinauer et al.}
%
\institute{Department of Neurology, Medical University of Graz, 8036 Graz, Austria
\email{\{christian.tinauer,christian.langkammer\}@medunigraz.at}\\
\url{http://www.neuroimaging.at} \and
Department of Radiology and Radiological Sciences, Monash University, 3800 Victoria, Australia \and
Fraunhofer Heinrich Hertz Institute, 10587 Berlin, Germany \and
Technische Universität Berlin, 10623 Berlin, Germany \and
BIFOLD – Berlin Institute for the Foundations of Learning and Data, 10587 Berlin, Germany}
\maketitle              
\begin{abstract}
\hfill\break
\textbf{\emph{Motivation.}} While recent studies show high accuracy in the classification of Alzheimer’s disease using deep neural networks, the underlying learned concepts have not been investigated.  

\textbf{\emph{Goals.}} To systematically identify changes in brain regions through concepts learned by the deep neural network for model validation.

\textbf{\emph{Approach.}} Using quantitative R2* maps we separated Alzheimer's patients (n=117) from normal controls (n=219) by using a convolutional neural network and systematically investigated the learned concepts using Concept Relevance Propagation and compared these results to a conventional region of interest-based analysis.

\textbf{\emph{Results.}} In line with established histological findings and the region of interest-based analyses, highly relevant concepts were primarily found in and adjacent to the basal ganglia.

\textbf{\emph{Impact.}} The identification of concepts learned by deep neural networks for disease classification enables validation of the models and could potentially improve reliability.

\end{abstract}
\section{Introduction}
Deep neural networks can learn features from magnetic resonance imaging (MRI) for the classification of Alzheimer's disease (AD) \cite{wen_convolutional_2020}, but are generally seen as black boxes \cite{davatzikos_machine_2019}. Medical applications are especially required to verify that the accuracy of those models is not the result of exploiting data artifacts. However, even when using conventional heat mapping approaches like Integrated Gradients \cite{sundararajan_axiomatic_nodate}, LIME \cite{ribeiro_why_2016} or Layer-wise Relevance Propagation (LRP) \cite{bach_pixel-wise_2015} it remains unclear whether atrophy, signal intensity changes or other contributors are relevant for the T1-based separation of patients with AD from normal controls (NC) \cite{tinauer_interpretable_2022}. Nevertheless, convolutional neural networks (CNNs) used for classification can be considered as feature extractors (e.g. channels of convolutional layers), which are later on classified by the fully connected layers. We make use of this network structure and the extracted features to explore the learned \emph{concepts} and their importance to the classification results.

In this work, we utilized a quantitative MRI parameter, the effective relaxation rate R2* for AD classification by a relevance-regularized CNN. The relaxation rate R2* is defined as the inverse of the effective transversal relaxation time T2* ($R_2^*=\frac{1}{T_2^*}$), whereas T2* can be considered the time observed for the transverse magnetization decay in each voxel.

R2* is highly positively correlated with iron concentration in gray matter \cite{langkammer_quantitative_2010}, and increased iron levels in the left and the right basal ganglia are frequent \emph{concepts} found in AD \cite{drayer_imaging_1988}. We believe that CNNs implicitly learn such concepts and have recently implemented the tools to disentangle and visualize them individually.

The goal of this paper is to investigate the usage of concepts learned by the CNN for AD classification using Concept Relevance Propagation (CRP) \cite{achtibat_attribution_2023}. Therefore, we compare \emph{differences in concept maps} with the outcome of an established region of interest (ROI) based analysis and relate the results to histologically well-known areas of disease activity.

\section{Methodology}
\subsection{Dataset}
We retrospectively selected 226 MRI datasets from 117 patients with probable AD (mean age=71.1±8.2 years, male/female=93/133) from our outpatient clinic and 226 MRIs from 219 propensity-logit-matched (covariates: age, sex) \cite{kline_psmpy_2022,rosenbaum_central_1983} normal controls (mean age=69.6±9.3 years, m/f=101/125) from an ongoing community-dwelling study. Patients and controls were scanned using a consistent MRI protocol at 3 Tesla (Siemens TimTrio) including a structural T1-weighted MPRAGE sequence with 1mm isotropic resolution (TR/TE/TI/FA = 1900 ms/2.19 ms/900 ms/9°, matrix = $176 \times 224 \times 256$) and a spoiled FLASH sequence (0.9x0.9x2mm³, TR/TE=35/4.92ms, 6 echoes, 4.92ms echo spacing, matrix = $208 \times 256 \times 64 \times 6$).

\subsection{Preprocessing}
Brain masks for each subject were obtained using FSL-SIENAX \cite{smith_accurate_2002} and the structural T1-weighted image. Using the data acquired from the spoiled FLASH sequence, we solved the inverse problem given as 

\begin{equation}
M_{xyz}[i] = M_{xyz}[0]e^{-t[i]R_2^*},
\end{equation}

for $M_{xyz}[0]$ and $R_2^*$, where $M_{xyz}[i]$ is the observed transversal magnetization in voxel $xyz$ at echo $i$ with echo time $t[i]$. The computations were executed for all voxels in the image volumes, yielding R2* maps (matrix = $208 \times 256 \times 64$).

The obtained R2* maps were affinely registered to the subject's MPRAGE sequence using FSL-flirt and nonlinearly registered to the MNI152 standard-space brain template using FSL-fnirt.

\subsection{Standard classification network}
We utilize a classifier network, which uses the combination of a single convolutional layer (kernel $3 \times 3 \times 3$, 8 channels) followed by a down-convolutional layer (kernel $3 \times 3 \times 3$, 8 channels, stridding $2 \times 2 \times 2$) as the main building block. The overall network stacks four of those main building blocks before passing the data through two fully connected layers (16 and 2 units). Each layer is followed by a Rectified Linear Unit (ReLU) nonlinearity, except for the output layer where a Softmax activation is applied. For better reproducibility, we initially created 3 initializations of the network weights and used them as starting points for the network trainings.

\subsection{Relevance-guided classification network}
To focus the network on relevant features, we proposed a relevance-guided network architecture, that extends the given classifier network with a relevance map generator \cite{tinauer_interpretable_2022}, named $Graz^+$. In brief, to guide the training process, we extended the classifier's categorical cross entropy loss ($\text{loss}_{\text{CCE}}$) by a loss term to act as a regularizer:

\begin{equation}
    \text{loss}_{\text{relevance}}(\mathbf{R}, \mathbf{M})=-\mathbf{1}^T\text{vec}(\mathbf{R} \odot \mathbf{M}),
\end{equation}

which enforces higher relevance values in $\mathbf{R}$ at positions marked by the focus region $\mathbf{M}$. Because the sum of relevance values in $\mathbf{R}$ is constant \cite{montavon_explaining_2017} this regularizer also automatically decreases the relevance values at positions not marked by $\mathbf{M}$. Consequently, this approach yields the total loss per data sample:

\begin{equation}
\begin{split}
    \text{loss}_{\text{Graz}^+}
    &= \text{loss}_{\text{relevance}} + \text{loss}_{\text{CCE}}\\
    &= -\mathbf{1}^T\text{vec}(\mathbf{R} \odot \mathbf{M}) -\sum_{i=1}^{\text{outputs}} y_i \cdot \log(\hat{y}_i),
\end{split}
\end{equation}

where $\mathbf{R}$ denotes the relevance heat map (3D shape), $\mathbf{M}$ is the predefined binary attention mask obtained during image preprocessing (3D shape), $\text{vec}(\mathbf{A})$ denotes the row major vector representation of $\mathbf{A}$ resulting in a column vector (1D shape), and $\mathbf{1}$ is a column vector where all elements are set to $1$ (1D shape). The inner product of the transposed vector $\mathbf{1}$ and the vector representation of $\mathbf{R} \odot \mathbf{M}$ gives the scalar value $\text{loss}_{\text{relevance}}$ (0D shape). The negative sign accounts for the maximization of the relevance inside the mask and $\odot$ denotes the element-wise product. For the categorical cross entropy $y_i$ is the target value of the $i$-th output class and $\hat{y}_i$ its predicted value. For this study we used the FSL-SIENAX brain masks from image preprocessing as attention masks. 

\subsection{Training}
We trained models on R2* maps in subject space using Adam \cite{kingma_adam_2015} for 60 epochs with a batch size of 6. AD and NC data were separately sampled into training, validation, and test sets (ratio 70:15:15) while maintaining all scans from one person in the same set. To ensure the same class distribution in all setups, final sets were created by combining one set from each cohort. The sampling procedure was repeated 10 times to enable a bootstrapping analysis.

We reduced the learning  by $0.3$ when the validation loss did not decrease for 5 consecutive epochs and we also reset the model's weights to the state at the begin of the plateau phase. Initial learning rate was set to $10^{-3}$ and minimum learning rate was set to $10^{-6}$.

\subsection{Model selection}
The optimal models were selected from the last epoch of each run. The relevance inside the attention mask (brain mask) threshold was set to 95\%, enforcing models where most of the relevance is inside the intracranial volume.

\subsection{Bootstrapping analysis}
For each data sampling (10 data samplings) and for all network weights initializations (3 weights initializations), we repeated training of the network, creating overall 30 training sessions with the same input image configuration \cite{bouthillier_accounting_2021}. To identify concepts, we selected the best-performing run in terms of validation accuracy from the 30 training sessions.

\subsection{Concepts identification}
\begin{figure}
\includegraphics[width=\textwidth]{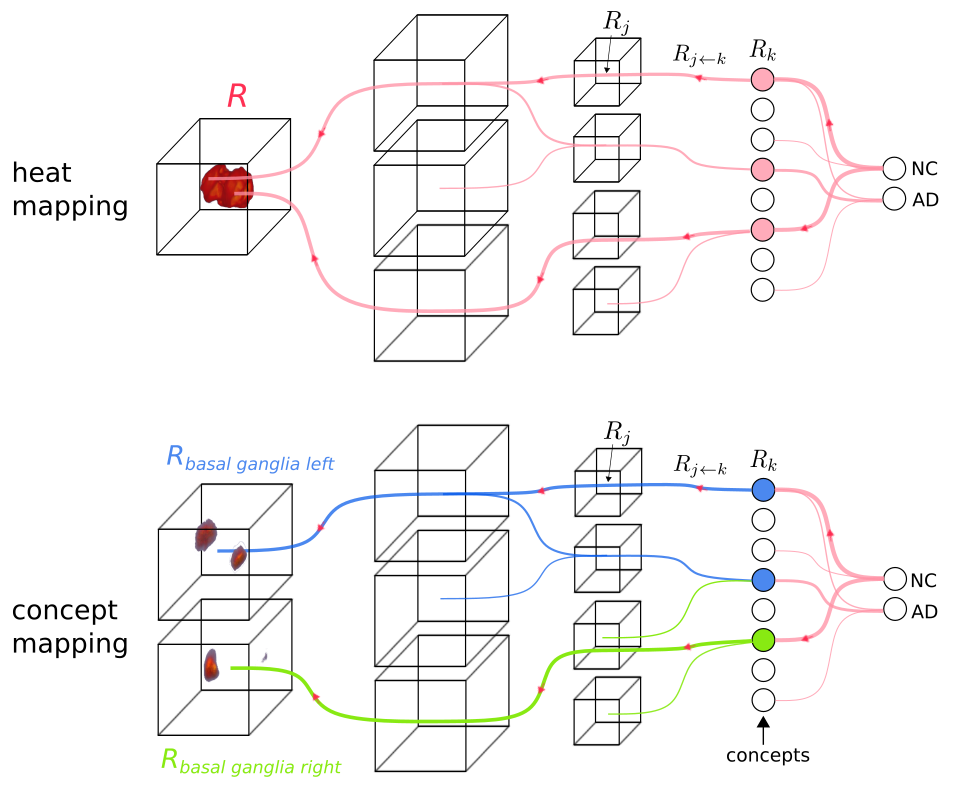}
\caption{Top row shows schematic overview of conventional back-propagation heat mapping propagating relevance scores backwards through the network creating a single heat map. Line thickness symbolizes the (relative) amount of relevance (in red) flowing through the connections. In contrast, shown in the bottom row, by conditioning on a concept encoded by a hidden-layer channel (highlighted in blue and green), Concept Relevance Propagation (CRP) and RelMax allow to compute concept-conditional explanations and provide semantic meaning for latent model structures, disentangling the learned and identified image features.
\textit{NC} normal control; \textit{AD} Alzheimer's disease; \textit{R} relevance.} \label{fig_crp_overview}
\end{figure}

Extending backpropagation-based heat mapping methods, CRP \cite{achtibat_attribution_2023} enables conditioning on a concept encoded by a hidden-layer channel. Hence, for this analysis we computed the concept-conditional explanations for all eight channels of the last down-convolutional layer before the fully connected layers (cf. Figure~\ref{fig_crp_overview} for details). All concept maps were computed w.r.t. both output classes and the $z^+$-composite applied. In this study, the applied $z^+$-composite is made up from the $z^+$-rule for all convolutional- and all fully connected layers. 

By applying RelMax \cite{achtibat_attribution_2023}, with the objective to maximize the relevance criterion, we ranked the test inputs for each concept channel. Furthermore, we computed the relative importance of the concept channels for the classification results and used it for ranking the concepts.

\subsection{Relevance-weighted concept map representation}
For local concept identification we present the top ranked test images in Figure~\ref{fig_heatmaps_top_ranked} in Appendix. Besides qualitatively investigating individual concept maps, we calculated mean concept maps for the classes NC and AD. As the concepts for both classes are visually very similar (cf. Figures~\ref{fig_heatmaps_most} and \ref{fig_heatmaps_least} in Appendix), we present the \emph{class differences of the concepts} given as

\begin{equation}
    \mathbf{R}_{diff} = \mathbf{R}_{NC,mean} - \mathbf{R}_{AD,mean},
\end{equation}

where $\mathbf{R}_{NC,mean}$ is the mean of the concept maps for the NC test images, and $\mathbf{R}_{AD,mean}$ is the mean of the concept maps for test AD images. Heat maps, concept maps, and differences of concept maps shown in this paper are overlaid on the MNI152 standard-space brain 1 mm template and windowed to present the top 50\% of relevance. Windowing maps to the top 50\% of relevance highlights voxel regions which receive high attributions of relevance, while suppressing less important regions. In mean maps five transversal slices are shown. In contrast, for individual concepts identification the same transversal slice for the five top ranked test images are presented.

\subsection{R2* and relevance region of interest analysis}
Anatomic structures were segmented on the T1-weighted MPRAGE sequence using FSL-FIRST \cite{patenaude_bayesian_2011}. Segmentation masks were applied to the R2* maps to calculate median R2* values in the basal ganglia (BG), hippocampi, and thalami \cite{damulina_cross-sectional_2020}. As median R2* values were not normally distributed, Mann-Whitney U tests were applied to study R2* differences between AD and NC. For concepts identification and comparison, the significance of relevance sum differences between AD and NC in the same brain regions were computed using Mann-Whitney U tests.

\section{Results}
The bootstrapping setup yielded a mean balanced accuracy and standard deviation of 75.64\%±5.16\%, sensitivity of 69.67\%±9.55\%, specificity of 81.61\%±5.45\% and an area under the curve of the receiver operating characteristics (AUC) of 0.76±0.05. 

\begin{figure}
\includegraphics[width=\textwidth]{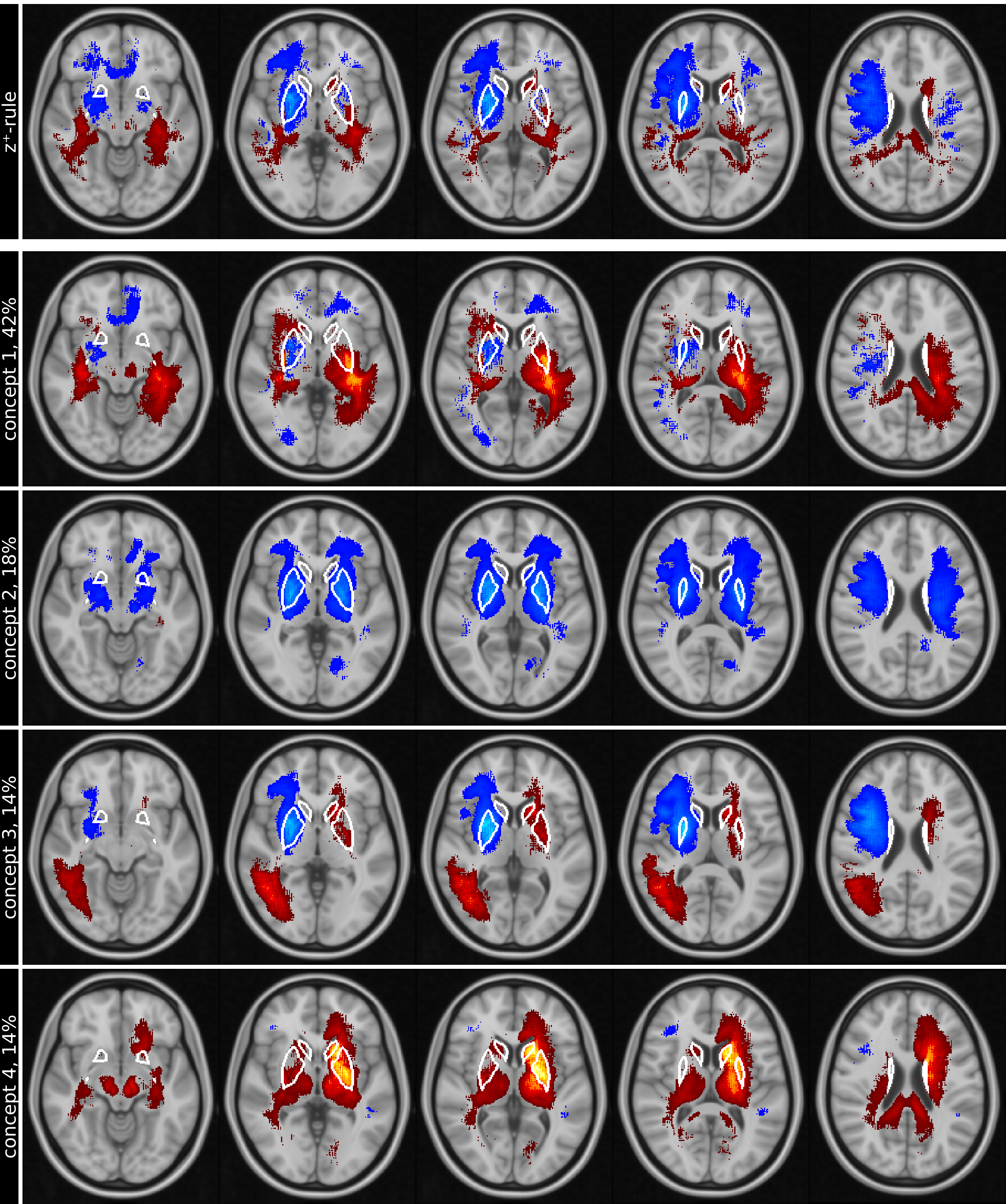}
\caption{Row (1) shows the difference of conventional mean global heat maps created for AD and NC using LRP-$z^+$-rule for five slices (columns), overlaid on the MNI152 standard-brain 1mm template. In comparison, rows (2) to (5) show the differences of the four most important concepts on the same slices, ranked by their relative importance (percentage next to name) for the classification results. Negative differences are presented as blue-lightblue and indicate regions in concepts with more attribution in AD compared to NC, whereas positive differences, shown as red-yellow, highlight regions in concepts with more attribution in NC compared to AD. If the attribution in NC and AD is similar (e.g. concept is nearly equally used for both classes) the regions disappear. Images are shown in standard-radiological view, thus left and right is flipped and white lines outline the basal ganglia.} \label{fig_heatmaps_diff}
\end{figure}

The differences between the concept attribution for AD and NC classification of the best-performing bootstrapping run are shown in Figure~\ref{fig_heatmaps_diff}. Additionally, Table~\ref{tab_relevances_sum_values} shows the relevance sum in the basal ganglia for the four most important concepts and the results of the tests for differences in the classes. The base images for computing the differences in concepts between AD and NC are shown in Figure~\ref{fig_heatmaps_most} and Figure~\ref{fig_heatmaps_least} in the Appendix.

Qualitatively, concept maps differences show a substantial overlap with the basal ganglia regions. The median R2* values in selected anatomical brain regions are given in Table~\ref{tab_r2star_median_values}.

\begin{table}
\caption{Sum of relevance values with interquartile range in parentheses in different brain regions in the four most important concepts. Mann-Whitney U test was used to calculate the $p$ Value for study participants with AD and NC. Basal ganglia is defined as the sum of caudate nucleus, pallidum, putamen according to \cite{khalil_determinants_2011}. * indicates that $p$ is less than $0.05$.}\label{tab_relevances_sum_values}
\begin{tabular}{|c|l|c|c|l|}
\hline
& {\bfseries Region} &  {\bfseries Relevance sum NC} & {\bfseries Relevance sum AD} & {\bfseries $p$ Value}\\
\hline
\hline
\multirow{8}{*}{{\rotatebox[origin=c]{90}{\bfseries concept 1}}}
& Left basal ganglia & 0.0129 (0.0106-0.0148) & 0.0077 (0.0060-0.0115) & {\bfseries 0.000*}\\
& Left caudate nucleus &  0.0020 (0.0016-0.0024) & 0.0014 (0.0011-0.0021) & {\bfseries 0.008*}\\
& Left pallidum & 0.0036 (0.0028-0.0041) & 0.0021 (0.0013-0.0030) & {\bfseries 0.001*}\\
& Left putamen & 0.0073 (0.0058-0.0082) & 0.0042 (0.0032-0.0064) & {\bfseries 0.000*}\\
\cline{2-5}
& Right basal ganglia & 0.0258 (0.0228-0.0291) & 0.0271 (0.0174-0.0312) & {\bfseries 0.974}\\
& Right caudate nucleus & 0.0032 (0.0023-0.0039) & 0.0031 (0.0026-0.0044) & {\bfseries 0.673}\\
& Right pallidum & 0.0061 (0.0051-0.0070) & 0.0064 (0.0044-0.0079) & {\bfseries 0.940}\\
& Right putamen & 0.0167 (0.0138-0.0185) & 0.0161 (0.0105-0.0193) & {\bfseries 0.650}\\
\hline
\hline
\multirow{8}{*}{{\rotatebox[origin=c]{90}{\bfseries concept 2}}}
& Left basal ganglia & 0.0025 (0.0019-0.0043) & 0.0072 (0.0047-0.0098) & {\bfseries 0.000*}\\
& Left caudate nucleus & 0.0004 (0.0003-0.0008) & 0.0012 (0.0008-0.0020) & {\bfseries 0.000*}\\
& Left pallidum & 0.0006 (0.0005-0.0011) & 0.0019 (0.0011-0.0024) & {\bfseries 0.000*}\\
& Left putamen & 0.0015 (0.0010-0.0025) & 0.0040 (0.0027-0.0055) & {\bfseries 0.000*}\\
\cline{2-5}
& Right basal ganglia & 0.0046 (0.0033-0.0057) & 0.0081 (0.0052-0.0119) & {\bfseries 0.000*}\\
& Right caudate nucleus & 0.0006 (0.0003-0.0007) & 0.0013 (0.0007-0.0017) & {\bfseries 0.000*}\\
& Right pallidum & 0.0013 (0.0007-0.0016) & 0.0022 (0.0014-0.0030) & {\bfseries 0.000*}\\
& Right putamen & 0.0030 (0.0020-0.0037) & 0.0051 (0.0034-0.0071) & {\bfseries 0.000*}\\
\hline
\hline
\multirow{8}{*}{{\rotatebox[origin=c]{90}{\bfseries concept 3}}}
& Left basal ganglia & 0.0062 (0.0046-0.0073) & 0.0040 (0.0025-0.0056) & {\bfseries 0.001*}\\
& Left caudate nucleus & 0.0015 (0.0011-0.0020) & 0.0010 (0.0006-0.0014) & {\bfseries 0.001*}\\
& Left pallidum & 0.0014 (0.0011-0.0018) & 0.0010 (0.0006-0.0013) & {\bfseries 0.004*}\\
& Left putamen & 0.0032 (0.0023-0.0039) & 0.0020 (0.0014-0.0029) & {\bfseries 0.002*}\\
\cline{2-5}
& Right basal ganglia & 0.0028 (0.0019-0.0043) & 0.0059 (0.0045-0.0083) & {\bfseries 0.000*}\\
& Right caudate nucleus & 0.0004 (0.0003-0.0006) & 0.0009 (0.0007-0.0013) & {\bfseries 0.000*}\\
& Right pallidum & 0.0007 (0.0005-0.0009) & 0.0012 (0.0010-0.0017) & {\bfseries 0.000*}\\
& Right putamen & 0.0029 (0.0025-0.0033) & 0.0022 (0.0016-0.0028) & {\bfseries 0.000*}\\
\hline
\hline
\multirow{8}{*}{{\rotatebox[origin=c]{90}{\bfseries concept 4}}}
& Left basal ganglia & 0.0120 (0.0085-0.0144) & 0.0068 (0.0034-0.0100) & {\bfseries 0.000*}\\
& Left caudate nucleus & 0.0029 (0.0021-0.0039) & 0.0013 (0.0007-0.0024) & {\bfseries 0.000*}\\
& Left pallidum & 0.0028 (0.0018-0.0035) & 0.0014 (0.0009-0.0024) & {\bfseries 0.000*}\\
& Left putamen & 0.0062 (0.0042-0.0070) & 0.0028 (0.0018-0.0049) & {\bfseries 0.000*}\\
\cline{2-5}
& Right basal ganglia & 0.0051 (0.0046-0.0060) & 0.0037 (0.0026-0.0048) & {\bfseries 0.000*}\\
& Right caudate nucleus & 0.0009 (0.0006-0.0013) & 0.0007 (0.0004-0.0009) & {\bfseries 0.012*}\\
& Right pallidum & 0.0013 (0.0011-0.0015) & 0.0009 (0.0005-0.0012) & {\bfseries 0.000*}\\
& Right putamen & 0.0016 (0.0012-0.0027) & 0.0036 (0.0028-0.0053) & {\bfseries 0.000*}\\
\hline
\end{tabular}
\end{table}

\begin{table}
\caption{Median R2* values with interquartile range in parentheses in different brain regions in study participants with AD and NC. Mann-Whitney U test was used to calculate the $p$ Value. Basal ganglia is defined as the mean of caudate nucleus, pallidum, putamen according to \cite{khalil_determinants_2011}. * indicates that $p$ is less than $0.05$.}\label{tab_r2star_median_values}
\begin{tabular}{|l|c|c|l|}
\hline
{\bfseries Region} &  {\bfseries Median $R_2^{*}$ NC ($sec^{-1}$)} & {\bfseries Median $R_2^{*}$ AD ($sec^{-1}$)} & {\bfseries $p$ Value}\\
\hline
\hline
Left basal ganglia &  29.00 (27.24-32.66) & 30.76 (28.15-33.60) & {\bfseries 0.006*}\\
Left caudate nucleus &  23.37 (21.43-25.70) & 24.28 (22.54-26.71) & {\bfseries 0.004*}\\
Left pallidum & 36.52 (33.17-40.58) & 36.43 (33.61-41.30) & {\bfseries 0.401}\\
Left putamen & 27.48 (25.15-31.75) & 30.11 (27.27-33.94) & {\bfseries 0.000*}\\
Left hippocampus & 16.40 (15.34-17.49) & 15.88 (14.78-17.06) & {\bfseries 0.003*}\\
Left thalamus & 19.70 (18.77-20.89) & 19.71 (18.85-20.92) & {\bfseries 0.972}\\
\hline
Right basal ganglia & 29.21 (26.74-32.14) & 30.62 (28.09-32.91) & {\bfseries 0.002*}\\
Right caudate nucleus & 23.06 (21.35-25.17) & 24.19 (22.11-25.94) & {\bfseries 0.002*}\\
Right pallidum & 36.25 (32.86-40.61) & 36.76 (33.68-41.54) & {\bfseries 0.094}\\
Right putamen & 28.08 (25.07-31.76) & 30.17 (27.08-34.11) & {\bfseries 0.000*}\\
Right hippocampus & 16.44 (15.54-17.38) & 16.02 (14.88-17.19) & {\bfseries 0.008*}\\
Right thalamus & 19.97 (18.82-20.71) & 19.76 (18.90-20.62) & {\bfseries 0.497}\\
\hline
\end{tabular}
\end{table}

\section{Discussion}
Previous studies using T1-weighted MRIs showed that volumetric brain features are highly relevant for the CNN-based classification of AD \cite{tinauer_interpretable_2022}, which has also been confirmed in recent related work on simulated aging \cite{hofmann_towards_2021}. To focus the classifier’s attention within brain parenchyma, we applied a relevance-guided regularized network ($Graz^+$) \cite{tinauer_interpretable_2022}. With the availability of accessible large MRI databases from patients, such as the Alzheimer’s Disease Neuroimaging Initiative (ADNI), AIBL or OASIS databases, various studies using CNN with explanation exploiting structural imaging data have been published \cite{wang_deep_2023,tang_interpretable_2019,bohle_layer-wise_2019,oh_classification_2019}. However, from T1-weighted images alone it cannot be concluded whether volumetric features, signal intensity changes or other factors are relevant.

Therefore, this study identified the concepts of the CNN-based classification with R2* maps. Various studies have shown that brain iron increasingly accumulates in the deep gray matter of AD patients \cite{damulina_cross-sectional_2020,lane_iron_2018}. While the highest relevances in and adjacent to the basal ganglia are consistently involved in all concepts, differences in concept mapping identified complementary spatial patterns (e.g. concept 1 and concept 3, Figure~\ref{fig_heatmaps_diff}).

The differences in concept 1 show substantial positive differences in and adjacent to the left basal ganglia, whereas the right basal ganglia show nearly no differences. This was confirmed by the tests for relevance sum differences in Table~\ref{tab_relevances_sum_values} and was also found in susceptibility sensitive imaging in AD \cite{damulina_cross-sectional_2020,acosta-cabronero_vivo_2013}. Concept 1 seems to activate more for NC images, favoring smaller R2* values in the left pallidum, left putamen and left white matter, whereas concept 2 shows only negative differences, favoring higher R2* values in the left and right basal ganglia. Additionally, concept 3 shows positive differences in the right temporal and occipital lobes, which is in line with the analysis in \cite{damulina_cross-sectional_2020}. Please note that all MR images in this paper are shown in standard-radiological view, causing the left and right side of the brain to be flipped.

Differences in the R2* values between AD and NC were additionally confirmed by the conventional ROI-based results in Table~\ref{tab_r2star_median_values}. 

\section{Conclusion}
This study used quantitative MRI data (R2*) for deep learning classification and CRP in a clinical cohort of AD patients. Confirming histological and in-vivo iron mapping studies, this underlines that heat- and concept mapping can serve as an exploratory means to identify areas of pathological tissue changes and further reveal internal mechanisms of deep learning classification networks. Using the differences in concept maps increases the spatial sparsity of relevances and might indicate which changes in the input images are learned by the CNNs within each concept.

\begin{credits}
\subsubsection{\ackname}
This study was funded by the Austrian Science Fund (FWF grant numbers: P30134,  P35887).
This work was supported by
the Federal Ministry of Education and Research (BMBF) as grant BIFOLD (01IS18025A, 01IS180371I);
the German Research Foundation (DFG) as research unit DeSBi (KI-FOR 5363);
the European Union’s Horizon Europe research and innovation programme (EU Horizon Europe) as grant TEMA (101093003);
the European Union’s Horizon 2020 research and innovation programme (EU Horizon 2020) as grant iToBoS (965221);
and the state of Berlin within the innovation support programme ProFIT (IBB) as grant BerDiBa (10174498).
This research was supported by NVIDIA GPU hardware grants.

\subsubsection{\discintname}
The authors have no competing interests to declare that are relevant to the content of this article.
\end{credits}

%
%
%
\bibliographystyle{splncs04}
\bibliography{xAI_2024}

\section*{Appendix}

\begin{figure}
\includegraphics[width=\textwidth]{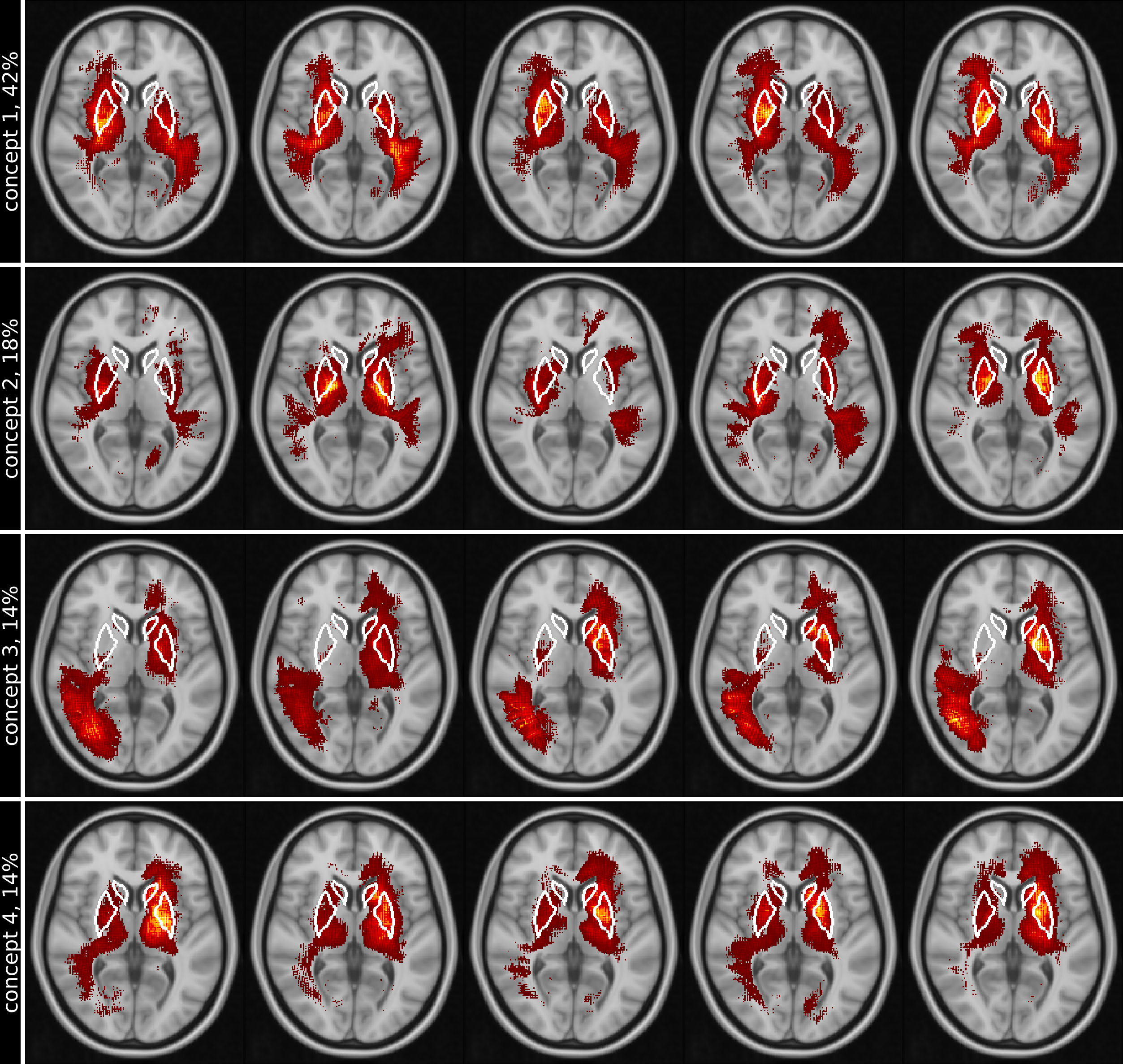}
\caption{Rows (1) to (4) show the 4 most important concepts, ranked by their relative importance (percentage next to name) for the classification results. Columns (1) to (5) show the 5 top ranked  individual test images for each concept on the same transversal slice. Positive relevance, shown as red-yellow, highlight regions in concepts with higher attributions. White lines in slices outline the basal ganglia. Images are shown in standard-radiological view, causing the left and right side of the brain to be flipped.} \label{fig_heatmaps_top_ranked}
\end{figure}

\begin{figure}
\includegraphics[width=\textwidth]{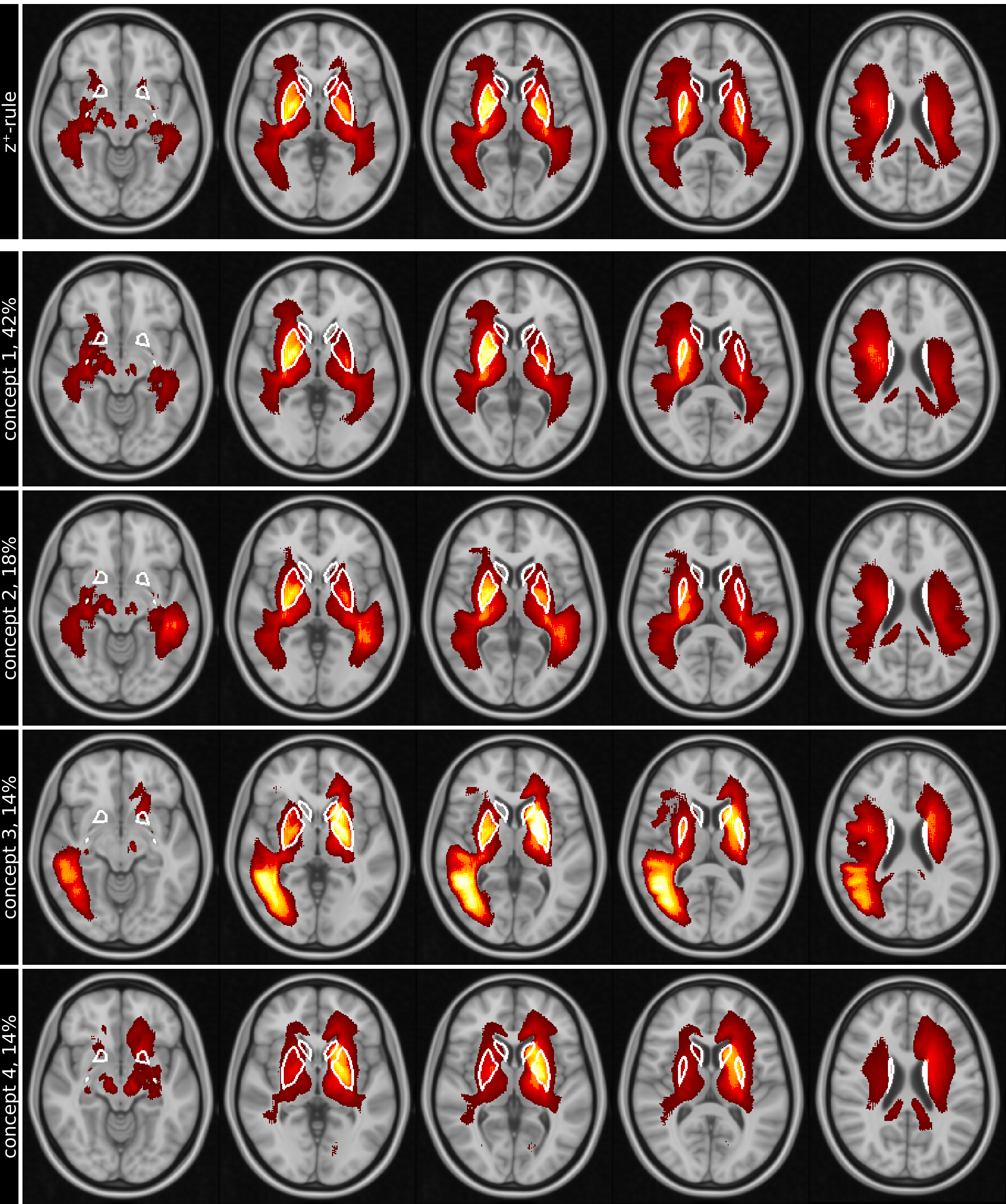}
\caption{Row (1) shows the conventional mean global heat maps created for NC using LRP-$z^+$-rule for 5 slices (columns), overlaid on the MNI152 standard-brain 1mm template. In comparison, rows (2) to (5) show the 4 most important concepts on the same slices, ranked by their relative importance (percentage next to name) for the classification results. Positive relevance, shown as red-yellow, highlight regions in concepts with higher attributions. White lines in slices outline the basal ganglia. Images are shown in standard-radiological view, causing the left and right side of the brain to be flipped.} \label{fig_heatmaps_most}
\end{figure}

\begin{figure}
\includegraphics[width=\textwidth]{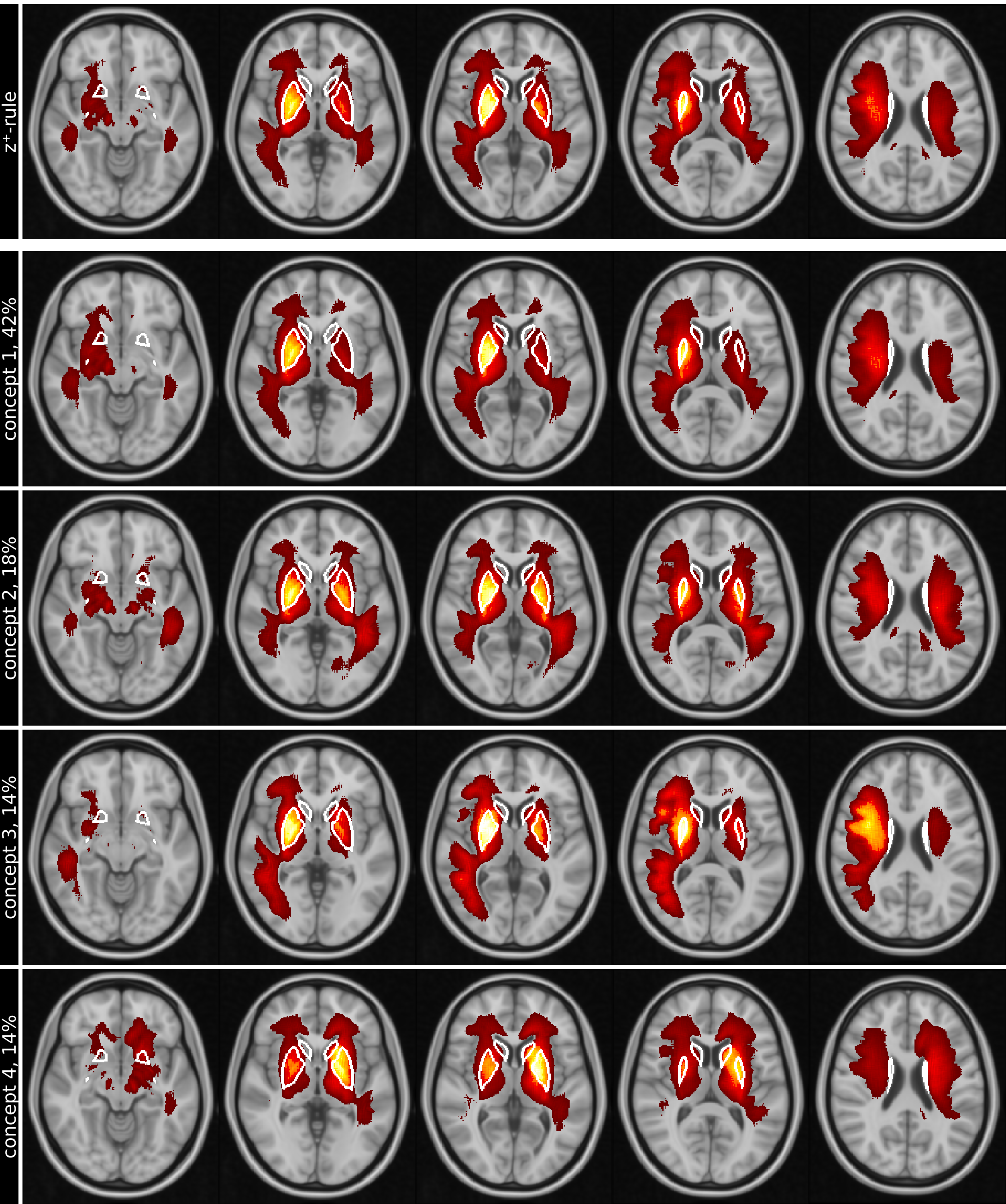}
\caption{Row (1) shows the conventional mean global heat maps created for AD using LRP-$z^+$-rule for 5 slices (columns), overlaid on the MNI152 standard-brain 1mm template. In comparison, rows (2) to (5) show the 4 most important concepts on the same slices, ranked by their relative importance (percentage next to name) for the classification results. Positive relevance, shown as red-yellow, highlight regions in concepts with higher attributions. White lines in slices outline the basal ganglia. Images are shown in standard-radiological view, causing the left and right side of the brain to be flipped.} \label{fig_heatmaps_least}
\end{figure}

Figure ~\ref{fig_heatmaps_top_ranked} shows the five top ranked test images for each concept on the same transversal slice, while figures~\ref{fig_heatmaps_most} and \ref{fig_heatmaps_least} show the mean concept maps for NC and AD on five transversal slices.

\end{document}